\documentclass[10pt,twocolumn,letterpaper]{article}

\usepackage{cvpr}              

\usepackage{amsmath}
\usepackage{algorithmic}
\usepackage{graphicx}
\usepackage{textcomp}
\usepackage{xspace}
\usepackage{amsfonts}
\usepackage{amssymb}
\usepackage{array}
\usepackage{xcolor}
\usepackage[table]{xcolor}
\usepackage{tabularx}
\usepackage{pifont}
\usepackage{multirow}
\usepackage{colortbl}
\usepackage{enumitem}
\usepackage{afterpage}
\usepackage[abbreviations]{glossaries-extra}

\definecolor{cvprblue}{rgb}{0.21,0.49,0.74}
\usepackage[pagebackref,breaklinks,colorlinks,allcolors=cvprblue]{hyperref}

\def\paperID{*****} 
\def\confName{CVPR}
\def\confYear{2026}

\title{Text-guided Fine-Grained Video Anomaly Understanding}

\author{
Jihao Gu$^{1}$ \quad
Kun Li$^{2}$ \quad
He Wang$^{1}$ \quad
Kaan Ak\c{s}it$^{1}$\\
$^{1}$University College London \\ 
$^{2}$CVLab, College of Information Technology, United Arab Emirates University\\
{\tt\small \{jihao.gu.23, he\_wang, k.aksit\}@ucl.ac.uk \quad
kunli.hfut@gmail.com}
}

\definecolor{ColorTVAU}{rgb}{0.8470588235294118, 0.10588235294117647, 0.3764705882352941}
\newcommand{\colorTVAU}[1]{\textcolor{ColorTVAU}{#1}}
\newabbreviation{TVAU}{T-VAU}{Text-guided Fine-Grained Video Anomaly Understanding}
\newcommand{\TVAU}{\textbf{\colorTVAU{\glsxtrshort{TVAU}}}\xspace}

\definecolor{ColorAHD}{RGB}{0,158,115}
\newcommand{\colorAHD}[1]{\textcolor{ColorAHD}{#1}}
\newabbreviation{AHD}{AHD}{Anomaly Heatmap Decoder}
\newcommand{\AHD}{\textbf{\colorAHD{\glsxtrshort{AHD}}}\xspace}

\definecolor{ColorRAE}{RGB}{230,159,0}
\newcommand{\colorRAE}[1]{\textcolor{ColorRAE}{#1}}
\newabbreviation{RAE}{RAE}{Region-aware Anomaly Encoder}
\newcommand{\RAE}{\textbf{\colorRAE{\glsxtrshort{RAE}}}\xspace}

\begin{document}
\maketitle
\begin{abstract}
  Subtle abnormal events in videos often manifest as weak spatio-temporal cues that are easily overlooked by conventional anomaly detection systems. Existing video anomaly detection approaches typically provide coarse binary anomaly decisions without interpretable evidence, while large vision-language models (LVLMs) can produce textual judgments but lack precise localization of subtle visual signals.
  To address this gap, we propose Text-guided Fine-Grained Video Anomaly Understanding (\TVAU), a framework that grounds subtle anomaly evidence into multimodal reasoning. Specifically, we introduce an Anomaly Heatmap Decoder (\AHD) that performs visual-textual feature alignment to extract pixel-level spatio-temporal anomaly heatmaps from intermediate visual representations. We further design a Region-aware Anomaly Encoder (\RAE) that converts these heatmaps into structured prompt embeddings, enabling the LVLM to perform anomaly detection, localization, and semantic explanation in a unified reasoning pipeline.
  To support fine-grained supervision, we construct a target-level fine-grained video–text anomaly dataset derived from ShanghaiTech and UBnormal with detailed annotations of object appearance, localization, and motion trajectories.
  Extensive experiments demonstrate that T-VAU significantly improves anomaly localization and textual reasoning performance on both benchmarks, achieving strong results in BLEU-4 metrics and Yes/No decision accuracy while providing interpretable pixel-level spatio-temporal evidence for anomaly understanding. The code will be available at \href{https://github.com/momiji-bit/T-VAU}{https://github.com/momiji-bit/T-VAU}.
\end{abstract}

\section{Introduction}
\label{sec:intro}

\begin{figure}[t]
  \centering
  \includegraphics[width=1\linewidth]{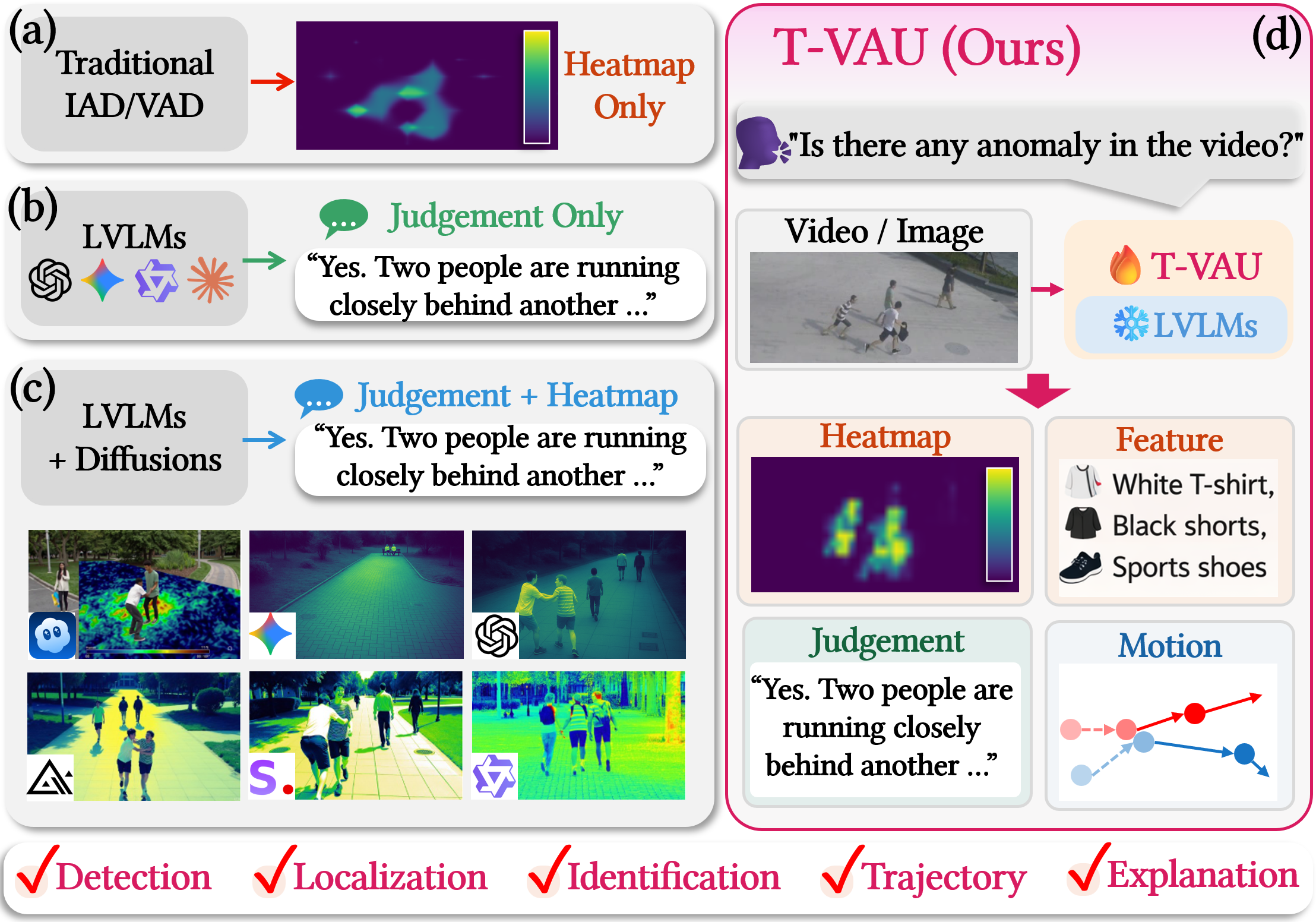}
  \caption{Comparison of anomaly detection paradigms and our \TVAU. (a) Traditional IAD/VAD outputs anomaly scores/heatmaps but lacks semantic explanation. (b) LVLMs provide textual judgments but lack pixel-level localization of subtle cues. (c) LVLM–diffusion hybrids combine visualization and text but may be unstable/inconsistent. (d) T-VAU couples a text-aligned \AHD with a \RAE to deliver pixel-level spatio-temporal localization and evidence-grounded reasoning (judgment, appearance, motion) for fine-grained anomaly understanding.}
  \label{fig:teaser}
  \vspace{-4pt}
\end{figure}

\afterpage{\afterpage{%
    \begin{figure*}[t!]
      \centering
      \includegraphics[width=\linewidth]{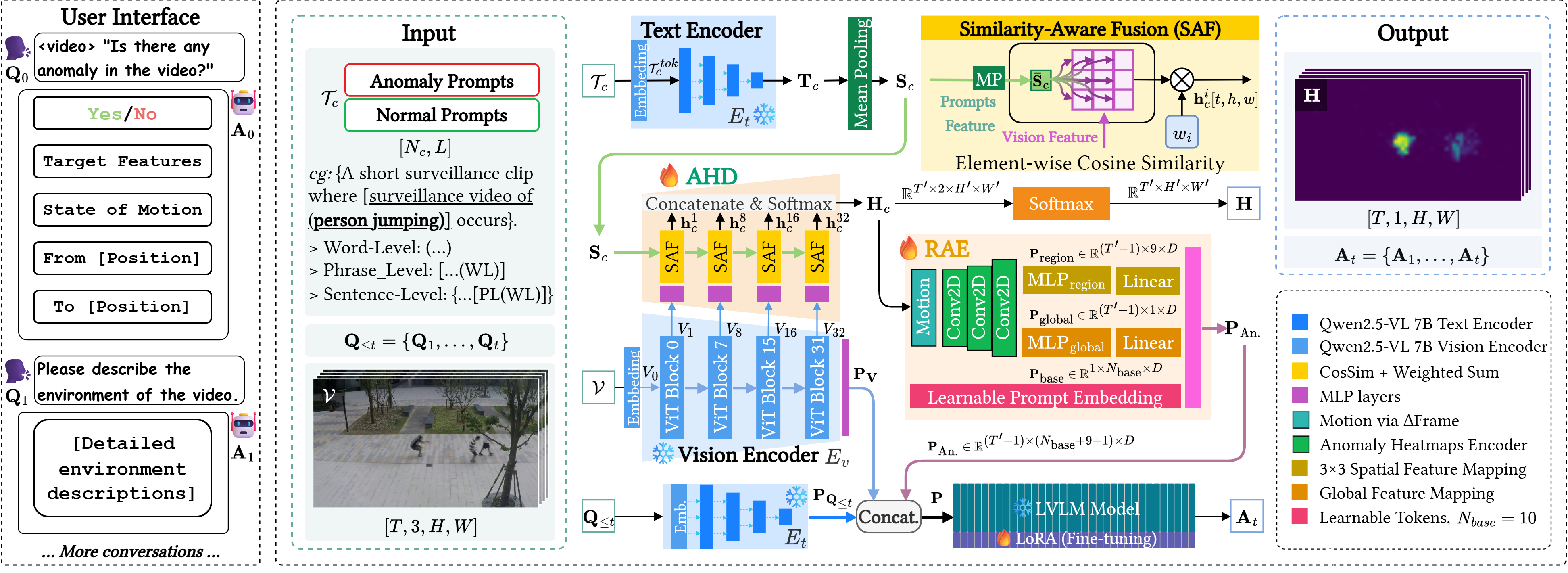}
      \caption{The proposed \TVAU model. The framework consists of three modules: a Text Encoder ($E_t$) that generates class-specific text embeddings $\mathbf{S}_c$ from binary prompts; an Anomaly Heatmap Decoder (\AHD) that fuses $\mathbf{S}_c$ with visual features $\mathcal{V}$ to produce spatio-temporal pixel-level anomaly heatmaps $\mathbf{H}$; and a Region-aware Anomaly Encoder (\RAE) that projects $\mathbf{H}_c$ into the LoRA-tuned LVLM semantic space and integrates it with a video $\mathcal{V}$ and a sequence of incrementally refined question $\mathbf{Q}_{\leq t}$ to yield the final anomaly understanding response $\mathbf{A}_t$.}
      \label{fig:framework}
      \vspace{-8pt}
    \end{figure*}
}}

Video Anomaly Detection (VAD)~\cite{sultani2018real,liu2018future,li2025anomize,wu2026deep,shao2025eventvad,ahn2026anyanomaly} is critical for safety-critical visual systems, \eg, public security monitoring and industrial inspection.
In real-world scenarios, anomalies often manifest as \emph{subtle spatio-temporal cues}—such as slight motion deviations, short-lived interactions, or fine-grained appearance changes—that are easily obscured by background clutter, compression artifacts, occlusions, and scene dynamics.
This low-SNR characteristic makes reliable detection highly dependent on \emph{interpretable pixel-level evidence}, a key aspect of subtle visual computing~\cite{lin2025svc,guo2023audio,guo2024benchmarking,yu2022physformer,gu2025motion,gu2025mm,qian2024dual}.

Mainstream VAD pipelines, including reconstruction-, prediction-, memory-based, and discriminative approaches, typically output video- or frame-level anomaly scores and rely on manual thresholds for decision-making~\cite{sultani2018real,liu2018future}.
Although effective in many cases, such coarse outputs are insufficient for subtle and localized anomalies: fine-grained cues may be diluted by feature aggregation, and the lack of explicit evidence reduces robustness, generalization, and human trust.
In practice, anomaly understanding often requires more than binary prediction, including \emph{where} the anomaly occurs, \emph{which} target is responsible, and \emph{how} it evolves over time.
Beyond detection, anomaly understanding therefore aims to localize abnormal events, identify responsible targets, and provide interpretable explanations of their temporal evolution, aligning with localization-oriented evaluation protocols, such as region-based and track-based criteria~\cite{ramachandra2020street}.

Recently, vision–language pretraining~\cite{radford2021clip,sun2023eva,liu2023visual,liu2024improved,li2022blip} and large vision–language models (LVLMs) have significantly advanced open-vocabulary perception and multimodal reasoning~\cite{bai2025qwen2,zhu2023minigpt,liu2024llavanext,chen2024internvl}.
These models are appealing for anomaly understanding because language provides a flexible interface for \emph{interactive querying} and \emph{semantic explanation}.
However, when directly applied to VAD, general-purpose LVLMs often struggle to \emph{ground} subtle anomalies in pixel-level evidence, leading to unreliable localization and potentially unfaithful textual claims (Fig.~\ref{fig:teaser}).
Meanwhile, text-guided anomaly localization has been explored in image anomaly understanding using VLP priors and fine-grained descriptions~\cite{gu2024filo}, but extending this capability to \emph{video} remains challenging due to temporal dynamics and subtle motion cues.

To address these limitations, we propose \TVAU, an LVLM-based framework that closes the loop from \emph{pixel-level evidence} to \emph{language-based reasoning}.
Our key idea is to (i) extract spatio-temporal anomaly evidence through visual–text alignment and (ii) inject this evidence into LVLMs as structured anomaly prompts for multi-task, multi-turn reasoning.
Specifically, we introduce an Anomaly Heatmap Decoder (\AHD) that aligns intermediate visual tokens with a \emph{built-in normal–abnormal prompt pair} to generate \emph{threshold-free} anomaly heatmaps with high spatial precision.
We further design a Region-aware Anomaly Encoder (\RAE) that converts heatmap evidence into region- and motion-aware prompt embeddings, bridging low-level anomaly evidence with high-level LVLM priors.
These components enable unified anomaly judgment, localization, target identification, appearance description, and motion or trajectory reasoning through multi-turn interaction.

To facilitate fine-grained anomaly understanding, we construct a target-level video–text anomaly dataset based on ShanghaiTech and UBnormal, with explicit supervision on \emph{appearance, localization, and motion trajectories}.
This dataset allows T-VAU to learn consistent mappings from pixel-level evidence to faithful language descriptions, improving both interpretability and generalization.

In summary, our contributions are threefold:

(1) We present \TVAU, a fine-grained video anomaly understanding framework built upon large vision–language models. It integrates an Anomaly Heatmap Decoder (\AHD) for precise pixel-level anomaly localization and a Region-aware Anomaly Encoder (\RAE) for injecting region-aware and motion-aware evidence into language reasoning, enabling a unified solution for anomaly detection, localization, and multi-turn explanation.

(2) We construct a fine-grained anomaly understanding dataset based on two large-scale benchmarks, ShanghaiTech and UBnormal, supporting both target localization and language-based anomaly understanding. It provides frame-wise annotations for ShanghaiTech (4,108 training / 1,028 validation) and target-aligned descriptions for ShanghaiTech (5,136 samples) and UBnormal (7,912 samples), covering abnormal targets, appearance attributes, and motion trajectories.

(3) We achieve state-of-the-art performance across anomaly judgment, localization, and dialog-based explanation benchmarks, demonstrating the effectiveness of our framework for fine-grained and explainable video anomaly understanding.

\section{Related Work}

\subsection{Video Anomaly Understanding}
Video anomaly detection (VAD) has long been dominated by the paradigm of learning normality and identifying deviations at test time, with representative directions including weakly supervised multiple-instance learning, predictive modeling, memory-based modeling, and more recent discriminative or density-based formulations~\cite{sultani2018real,liu2018future,zhong2019gcn,park2020memory,wu2024mulde}. While these methods have substantially advanced benchmark performance, most of them primarily output frame- or clip-level anomaly scores, making it difficult to support fine-grained reasoning about \emph{where} an anomaly occurs, \emph{which} target is responsible, and \emph{how} it evolves over time. This limitation has motivated a shift toward localization-oriented evaluation and finer anomaly localization, as reflected by region- and track-based protocols~\cite{ramachandra2020street} and benchmarks with pixel-level annotations such as UBnormal~\cite{acsintoae2022ubnormal}. In parallel, vision-language pretraining has introduced a new line of work that injects semantic priors into VAD, including prompt-based weakly supervised methods such as VadCLIP and STPrompt~\cite{wu2024vadclip,wu2024stprompt}, open-vocabulary formulations such as OVVAD~\cite{wu2024ovvad}, and training-free or reasoning-oriented frameworks built upon large language models, \eg, LAVAD and AnomalyRuler~\cite{zanella2024lavad,yang2024anomalyruler}. 

More recently, video anomaly understanding (VAU) has emerged as a distinct setting, with datasets and systems such as UCA, CUVA, HAWK, VERA, Holmes-VAU, VANE-Bench, and VAU-R1 pushing the field from score prediction toward explanation, dialogue, and anomaly-aware reasoning~\cite{yuan2024uca,du2024cuva,tang2024hawk,ye2025vera,zhang2025holmesvau,bharadwaj2025vanebench,zhu2025vaur1}. Nevertheless, existing approaches still often rely on coarse global descriptions or weak spatial grounding, leaving a gap between pixel-level anomaly evidence and faithful language-based interpretation.

\subsection{Subtle Visual Cues and Fine-grained Supervision}
A key bottleneck in fine-grained anomaly understanding lies in the mismatch between the desired level of reasoning and the granularity of available supervision. Classical VAD datasets, such as ShanghaiTech and UCF-Crime, mainly provide video- or frame-level labels, which are sufficient for anomaly scoring but inadequate for learning consistent mappings from localized evidence to target-aware semantic explanations~\cite{liu2018future,sultani2018real}. Although subsequent benchmarks have gradually enriched the supervision space, their focus varies substantially: Street Scene emphasizes localization-oriented evaluation~\cite{ramachandra2020street}, UBnormal provides pixel-level masks under an open-set setting~\cite{acsintoae2022ubnormal}, UCA extends surveillance anomaly understanding with sentence-level annotations~\cite{yuan2024uca}, and CUVA further introduces causal supervision over the \emph{what}, \emph{why}, and \emph{how} of anomalous events~\cite{du2024cuva}. Recent large-scale resources, including Holmes-VAU, SurveillanceVQA-589K, and VANE-Bench, continue this trend by expanding anomaly understanding toward long-horizon, hierarchical, and conversational settings~\cite{zhang2025holmesvau,liu2025surveillancevqa,bharadwaj2025vanebench}. These developments are also closely aligned with the agenda of subtle visual computing (SVC)~\cite{lin2025svc,li2023joint,liu2024micro,guo2024ma,li2025mmad,li2025prototypical,li2026mabench,qian2024cluster,guo2024mac,li2025mac}, which emphasizes low-SNR visual cues, robustness, generalization, interpretability, and unified modeling across tasks and modalities in complex real-world environments~\cite{svc2026}. From this perspective, fine-grained video anomaly understanding can be viewed as an instance of SVC: subtle anomalies are often local, short-lived, and easily drowned by background clutter, requiring not only stronger spatio-temporal representations but also evidence-aware explanation. 

Our work follows this direction by coupling pixel-level anomaly grounding with target-level appearance, localization, and trajectory supervision, thereby enabling anomaly judgment, localization, and language explanation within a unified framework.

\section{Proposed Method}
\label{sec:method}

\subsection{Framework Overview}
We propose \TVAU, a unified framework for fine-grained video anomaly understanding built on a frozen Large Vision-Language Model (LVLM). Unlike conventional video anomaly detection methods that mainly output anomaly scores or coarse localization results, T-VAU jointly supports anomaly judgment, spatio-temporal localization, and interactive language-based analysis. Given an input video $\mathcal{V} \in \mathbb{R}^{T \times 3 \times H \times W}$, a sequence of natural language queries $\mathbf{Q}_{\leq t}=\{\mathbf{Q}_1,\ldots,\mathbf{Q}_t\}$, and binary text prompts $\mathcal{T}_c \in \mathbb{R}^{N_c \times L}$ for each category $c \in \{\mathrm{normal}, \mathrm{abnormal}\}$, the model predicts pixel-level spatio-temporal anomaly heatmaps $\mathbf{H}$ and the response $\mathbf{A}_t$ at dialogue round $t$:
\begin{equation}
  \mathbf{H}, \mathbf{A}_t = \mathcal{M}(\mathcal{V}, \mathbf{Q}_{\leq t}, \mathbf{A}_{\leq t-1}, \mathcal{T}_c; \Theta),
\end{equation}
where $\mathcal{M}(\cdot;\Theta)$ denotes \TVAU, and $T'$, $H'$, and $W'$ denote the temporal and spatial resolutions after the visual encoder.

As shown in Figure~\ref{fig:framework}, T-VAU consists of a frozen LVLM backbone, including a Vision Encoder $E_v$, a Text Encoder $E_t$, and a Large Language Model Decoder $D_l$, together with two lightweight trainable modules: the AHD and the RAE. The \AHD aligns multiscale visual features with sentence-level text embeddings to produce frame-wise anomaly heatmaps. The \RAE further converts these heatmaps into structured prompt embeddings, which are fused with visual and dialogue context to enhance anomaly-aware reasoning. In this way, T-VAU couples explicit anomaly localization with language-guided reasoning in a unified framework.

\subsection{Model Design}
\subsubsection{Anomaly Heatmap Decoder}

\AHD takes multiscale visual features from $E_v$ and binary text prompts from $E_t$ as input. For each scale, it computes visual-text cosine similarity and fuses the resulting similarity maps to generate anomaly heatmaps. During this stage, only AHD is optimized.

\textbf{Textual Feature Extraction.}
We construct template-based prompts for two categories, \emph{normal} and \emph{abnormal}, and tokenize them as $\mathcal{T}^{\mathrm{tok}}_c \in \mathbb{R}^{N_c \times L}$. The text encoder $E_t$ extracts token-level embeddings, which are mean-pooled to obtain sentence-level features:
\begin{equation}
  \mathbf{T}'_c = E_t(\mathcal{T}^{\mathrm{tok}}_c) \in \mathbb{R}^{N_c \times L \times D},
\end{equation}
\begin{equation}
  \mathbf{T}_c = \frac{1}{N_c L}
  \sum_{n=1}^{N_c}\sum_{\ell=1}^{L} \mathbf{T}'_c[n,\ell,:] \in \mathbb{R}^{D},
\end{equation}
where $D$ is the text feature dimension.

\textbf{Visual Feature Extraction.}
Given a video clip $\mathcal{V} \in \mathbb{R}^{T \times 3 \times H \times W}$,
we first encode it using the visual encoder to obtain an initial feature tensor
$\mathbf{V}_0 \in \mathbb{R}^{T' \times D_v \times H' \times W'}$.
Let $\phi_i(\cdot)$ denote the output of the $i$-th transformer block in the visual encoder.
We extract intermediate visual representations from multiple layers to capture multi-scale semantics:
\begin{equation}
  \mathbf{V}_{i} = \phi_i(\mathbf{V}_{0}), \quad i \in \{1,8,16,32\},
\end{equation}
where $\mathbf{V}_{i} \in \mathbb{R}^{T' \times D_v \times H' \times W'}$ denotes the feature map from the $i$-th transformer layer.
These features encode visual information at different levels of abstraction and are later fused to construct anomaly heatmaps.

\textbf{Anomaly Heatmaps Generation.}
We project each intermediate visual feature $\mathbf{V}_{i}$ to the shared text-visual space via an MLP, yielding
$\mathbf{V}'_{i} \in \mathbb{R}^{T' \times D \times H' \times W'}$.
We then compute cosine similarity between each spatio-temporal visual token and the sentence-level text feature:
\begin{equation}
  \mathbf{h}^{i}_{c}[t, h, w] =
  \mathrm{CosineSimilarity}\!\left(\mathbf{V}'_{i}[t, :, h, w], \mathbf{T}_{c}\right).
\end{equation}
To aggregate information across layers, we introduce learnable weights $w_i$ and fuse the similarity maps:
\begin{equation}
  \mathbf{H}_c = \sum_{i} w_{i} \cdot \mathbf{h}^{i}_{c}
  \in \mathbb{R}^{T' \times 2 \times H' \times W'}.
\end{equation}
Finally, we apply softmax along the category dimension and take the anomaly channel as the final heatmap:
\begin{equation}
  \tilde{\mathbf{H}} = \mathrm{Softmax}(\mathbf{H}_c)
  \in \mathbb{R}^{T' \times 2 \times H' \times W'},
\end{equation}
\begin{equation}
  \mathbf{H} = \tilde{\mathbf{H}}[:,\,\mathrm{abnormal},:,:]
  \in \mathbb{R}^{T' \times 1 \times H' \times W'}.
\end{equation}


\subsubsection{Region-aware Anomaly Encoder}

\RAE transforms predicted anomaly heatmaps into structured prompt embeddings, which are injected into the LVLM decoder to improve anomaly reasoning.

\textbf{Region-Aware Feature Extraction.}
Given category-wise anomaly heatmaps sequence $\mathbf{H}_c \in \mathbb{R}^{T' \times 2 \times H' \times W'}$, we first compute temporal differences between adjacent frames:
\begin{equation}
  \mathbf{X}[t] = \mathbf{H}_c[t+1] - \mathbf{H}_c[t],
  \quad \{t = 1,\ldots,T'-1\},
\end{equation}
where $\mathbf{X} \in \mathbb{R}^{(T'-1) \times 2 \times H' \times W'}$
represents the temporal differences between consecutive anomaly heatmaps.
These motion-aware heatmaps are then processed by a lightweight convolutional backbone to extract region-aware features:
\begin{equation}
  \mathbf{F} = \mathrm{ConvBackbone}(\mathbf{X}) \in \mathbb{R}^{(T'-1) \times C_h \times H' \times W'}.
\end{equation}

\textbf{Prompt Embedding Generation.}
To encode both local and global anomaly cues, we partition each frame into a regular grid (e.g., $3\times3$) and apply adaptive average pooling to obtain regional features $\mathbf{F}_{\mathrm{grid}} \in \mathbb{R}^{(T'-1) \times 9 \times C_h}$. These features are mapped into region-specific prompts:
\begin{equation}
  \mathbf{P}_{\mathrm{region}} = \mathrm{MLP}_{\mathrm{region}}(\mathbf{F}_{\mathrm{grid}})
  \in \mathbb{R}^{(T'-1) \times 9 \times D}.
\end{equation}
A global prompt is generated by spatial mean pooling followed by an MLP:
\begin{equation}
  \mathbf{p}_{\mathrm{global}} = \mathrm{MLP}_{\mathrm{global}}(\mathrm{MeanPool}(\mathbf{F}))
  \in \mathbb{R}^{(T'-1) \times 1 \times D}.
\end{equation}
We also introduce a learnable base prompt $\mathbf{P}_{\mathrm{base}} \in \mathbb{R}^{1 \times N_{\mathrm{base}} \times D}$, shared across time. The final anomaly prompt sequence is constructed as
\begin{equation}
  \mathbf{P}_{\mathrm{An}} =
  \big[\,\mathbf{P}_{\mathrm{base}},\ \mathbf{P}_{\mathrm{region}},\ \mathbf{p}_{\mathrm{global}}\,\big]
  \in \mathbb{R}^{(T'-1) \times (N_{\mathrm{base}}+9+1) \times D}.
\end{equation}

We then concatenate anomaly prompts with visual prompts and dialogue context:
\begin{equation}
  \begin{aligned}
    \mathbf{P}_{\mathbf{V}} &= \mathrm{MLP}(\mathbf{V}_{32}), \\
    \mathbf{P}_{\mathbf{Q}_{\leq t}} &= E_t(\mathbf{Q}_{\leq t}, \mathbf{A}_{\leq t-1}), \\
    \mathbf{P} &= \big[\,\mathbf{P}_{\mathbf{V}},\ \mathbf{P}_{\mathrm{An}},\ \mathbf{P}_{\mathbf{Q}_{\leq t}}\,\big],
  \end{aligned}
\end{equation}
where $\mathbf{P}_{\mathbf{V}}$ is the visual prompt, $\mathbf{P}_{\mathbf{Q}_{\leq t}}$ encodes the query-response history, and $\mathbf{P}$ is the final prompt sequence fed into $D_l$ for anomaly-aware reasoning.

\begin{figure*}[t!]
  \centering
  \includegraphics[width=\linewidth]{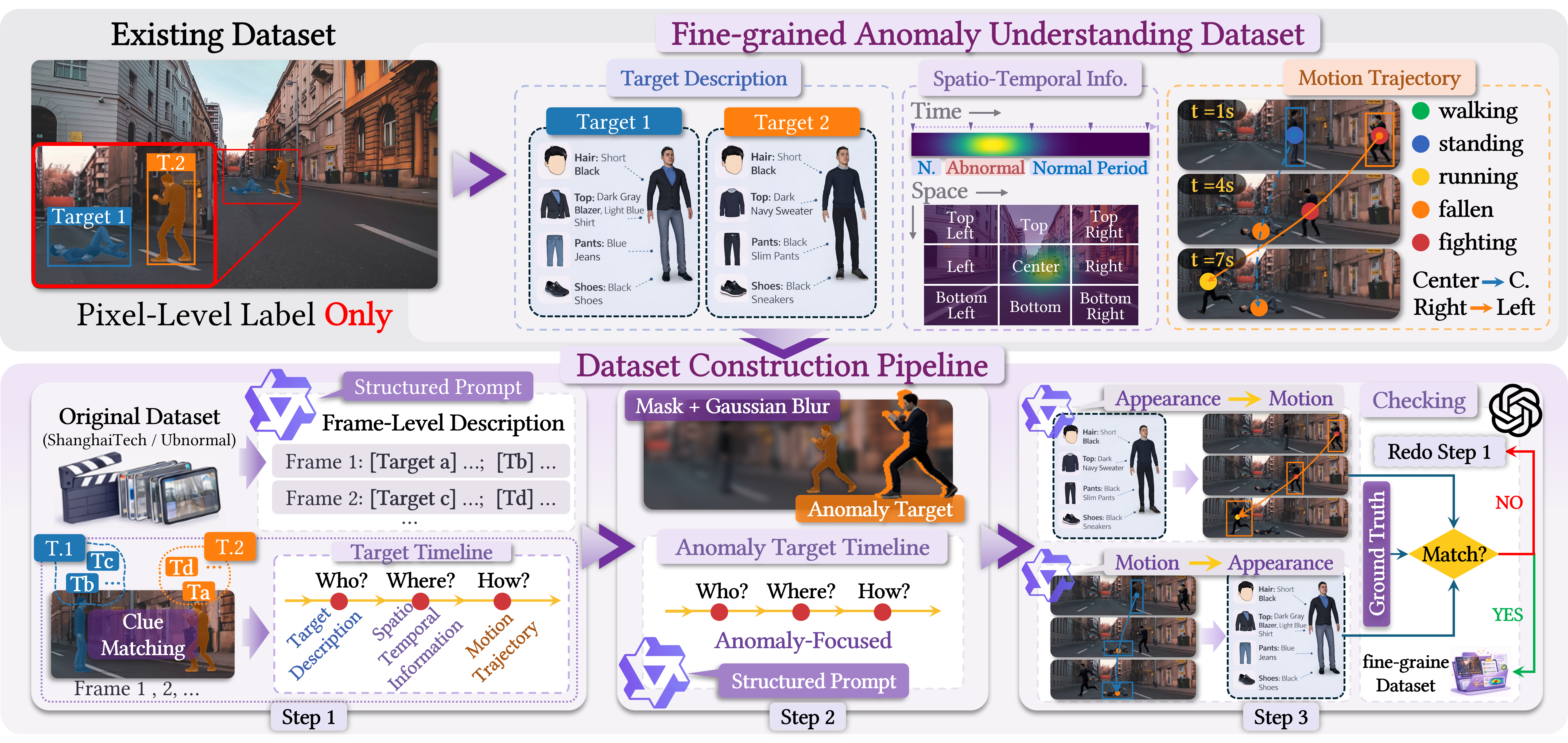}
  \caption{Fine-grained anomaly dataset construction pipeline.
    Starting from existing datasets with only pixel-level anomaly labels (e.g., ShanghaiTech and UBnormal),
    we build a structured video–text dataset through three stages:
    (1) frame-level structured prompting to extract target attributes and spatial information and aggregate them into target timelines;
    (2) anomaly-focused refinement using anomaly masks and background suppression to emphasize abnormal evidence;
    and (3) cross-modal consistency verification between appearance and motion cues.
  The resulting dataset provides aligned annotations of appearance, spatial localization, and motion trajectory for fine-grained anomaly understanding.}
  \label{fig:dataset}
  \vspace{-8pt}
\end{figure*}

\subsection{Fine-grained Anomaly Understanding} \label{sec:dataset}

To support fine-grained anomaly understanding, we construct a structured video–text supervision pipeline covering three aspects: appearance attributes, spatial localization, and motion trajectory, as illustrated in Fig.~\ref{fig:dataset}. Since directly applying LVLMs to videos often loses fine-grained spatio-temporal details, we adopt a frame-wise extraction and temporal aggregation strategy to align visual evidence with textual descriptions.

\textbf{Frame-level extraction and temporal aggregation.}
Each frame is processed with a unified structured prompt to extract target-level attributes and bounding boxes. The frame-wise outputs are then linked across time through identity association and spatial consistency, forming target-level timelines that describe who appears, where the event occurs, and how it evolves.

\textbf{Anomaly-focused refinement.}
To emphasize anomalous evidence, we further refine the data by applying anomaly masks and suppressing background regions with Gaussian blurring. We then re-extract target-level descriptions from the refined frames and aggregate them into high-confidence anomaly timelines, providing stronger supervision for anomaly localization and motion reasoning.

\textbf{Cross-modal consistency verification.}
Finally, we introduce bidirectional verification between appearance and motion cues. In the appearance $\rightarrow$ motion path, motion trajectories are inferred from appearance-conditioned timelines; in the motion $\rightarrow$ appearance path, motion patterns are used to verify appearance-based descriptions. This process enforces consistency between localization, trajectory, and textual semantics, and aligns the supervision signals with pixel-level spatio-temporal anomaly heatmaps generated by \AHD.

Following this pipeline, we obtain a structured anomaly supervision dataset with aligned annotations of appearance, localization, and temporal trajectory, which provides fine-grained supervision for evidence-grounded anomaly understanding.

\section{Experiments}

\subsection{Experimental Setup}

\textbf{Datasets.} \textbf{UBnormal}~\cite{acsintoae2022ubnormal} is a synthetic dataset for supervised open-set anomaly detection. It contains 543 clips (236,902 frames) generated via Cinema4D across 29 virtual scenes, with pixel-level target annotations. The dataset covers 22 anomaly types (\eg, fighting and fire), which are mutually exclusive across splits: training contains both normal and abnormal events, whereas testing includes disjoint abnormal categories. \textbf{ShanghaiTech}~\cite{liu2018future} is a widely used unsupervised benchmark collected in real campus environments. It consists of 330 normal training videos and 107 testing videos (including both normal and abnormal events) across 13 scenes. The test set contains 130 annotated anomalies, such as vehicles in pedestrian areas, fighting, and cycling.
We use the augmented datasets derived from UBnormal and ShanghaiTech as training and validation data, as described in Section~\ref{sec:dataset}.

\textbf{Evaluation Metrics for \AHD.} We use frame-level Area Under the Curve (AUC)~\cite{huang2005using}, reporting both micro and macro versions following~\cite{georgescu2021background}. We also adopt the Region-Based (RBDC) and Track-Based (TBDC) Detection Criteria~\cite{ramachandra2020street} to evaluate spatial localization.

\textbf{Evaluation Metrics for \RAE.} We evaluate \RAE on \emph{anomaly explanation} and \emph{discriminative ability}. For faithfulness, we compute BLEU-4~\cite{papineni2002bleu} ($\mathrm{BP} \cdot \exp(\tfrac{1}{4}\sum \log p_n)$) separately for \emph{Target} and \emph{Trajectory} descriptions, using the best-matching reference per video. For discriminative ability, we report binary accuracy ($\mathrm{Acc}$) and balanced accuracy ($\mathrm{bAcc} = \tfrac{1}{2}(\mathrm{TPR} + \mathrm{TNR})$). We further evaluate threshold-free discrimination using frame-level micro- and macro-ROC–AUC. All experiments adopt a one-shot setting with minimal text normalization.


\begin{table}[t!]
  \centering
  \resizebox{\columnwidth}{!}{%
    \begin{tabular}{|l|cc|c|c|}
      \hline
      \multirow{3}{*}{\textbf{Method}}
      & \multicolumn{4}{c|}{\textbf{Validation}} \\
      \cline{2-5}
      & \multicolumn{2}{c|}{AUC} & \multirow{2}{*}{RBDC $\uparrow$} & \multirow{2}{*}{TBDC $\uparrow$} \\
      \cline{2-3}
      & Micro $\uparrow$ & Macro $\uparrow$ & & \\
      \hline
      Georgescu \etal~\cite{georgescu2021background} & 58.5 & \cellcolor{green!20}94.4 & 18.580 & 48.213 \\
      Georgescu \etal~\cite{georgescu2021background} (FT)
      & 68.2 & \cellcolor{green!45}\textbf{95.3} & 28.654 & 58.097 \\
      \hline
      Sultani \etal~\cite{sultani2018real} (PT) & 61.1 & 89.4 & 0.001 & 0.012 \\
      Sultani \etal~\cite{sultani2018real} (FT) & 51.8 & 88.0 & 0.001 & 0.001 \\
      \hline
      Bertasius \etal~\cite{bertasius2021space} (FT, SR=1/32)
      & 86.1 & 89.2 & 0.008 & 0.021 \\
      Bertasius \etal~\cite{bertasius2021space} (FT, SR=1/8)
      & 83.4 & 90.6 & 0.009 & 0.023 \\
      Bertasius \etal~\cite{bertasius2021space} (FT, SR=1/4)
      & 78.5 & 89.2 & 0.006 & 0.018 \\
      \hline
      \AHD (1S)
      & \cellcolor{green!20}94.5 & 85.2 & \cellcolor{green!20}64.3 & \cellcolor{green!20}74.4 \\
      \AHD (FT)
      & \cellcolor{green!45}\textbf{94.8} & 87.8 & \cellcolor{green!45}\textbf{67.8} & \cellcolor{green!45}\textbf{76.7} \\
      \hline
    \end{tabular}%
  }
  \caption{
  Experimental results on UBnormal. We report micro-/macro-averaged frame-level AUC, RBDC, and TBDC (\%) for the baselines~\cite{georgescu2021background,sultani2018real,bertasius2021space}. PT, FT, and 1S denote pre-trained, fine-tuned, and one-shot; SR denotes frame sampling rate. Although only~\cite{georgescu2021background} supports anomaly localization, we report RBDC and TBDC for all baselines for completeness. Best results are highlighted in bold with a green background.}
  \label{tab:val_test_results}
  \vspace{-4pt}
\end{table}

\begin{table*}[t!]
  \centering
  \resizebox{\textwidth}{!}{
    \begin{tabular}{|l|c|ccc|ccc|}
      \hline
      \multirow{3}{*}{\textbf{Method}} &
      \multirow{3}{*}{\textbf{Size $\downarrow$}} &
      \multicolumn{3}{c|}{\textbf{ShanghaiTech}} &
      \multicolumn{3}{c|}{\textbf{UBnormal}} \\
      \cline{3-8}
      & & \multicolumn{2}{c|}{\textbf{BLEU-4 $\pm$ Std $\uparrow$}} & \textbf{Acc. $\pm$ Std $\uparrow$} &
      \multicolumn{2}{c|}{\textbf{BLEU-4 $\pm$ Std $\uparrow$}} & \textbf{Acc. $\pm$ Std $\uparrow$} \\
      \cline{3-8}
      & & \textbf{Target} & \textbf{Trajectory} & \textbf{Yes/No} &
      \textbf{Target} & \textbf{Trajectory} & \textbf{Yes/No} \\
      \hline\hline
      Qwen2.5-VL (zero-shot) & 7B & 18.74 $\pm$ 0.82 & 27.33 $\pm$ 1.05 & 61.03 $\pm$ 0.38\% & 16.20 $\pm$ 0.85 & 24.18 $\pm$ 1.08 & 65.62 $\pm$ 0.41\% \\
      Qwen2.5-VL (one-shot) & 7B & 50.42 $\pm$ 0.58 & 78.91 $\pm$ 0.72 & 92.36 $\pm$ 0.24\% & 44.35 $\pm$ 0.62 & 70.82 $\pm$ 0.75 & 87.24 $\pm$ 0.27\% \\
      LLaVA-1.6 (one-shot) & 7B & 47.68 $\pm$ 0.60 & 75.42 $\pm$ 0.74 & 91.07 $\pm$ 0.26\% & 42.11 $\pm$ 0.64 & 68.07 $\pm$ 0.78 & 85.91 $\pm$ 0.28\% \\
      MiniCPM-V 2.6 (one-shot) & 7B & 52.34 $\pm$ 0.54 & 80.41 $\pm$ 0.69 & 93.11 $\pm$ 0.23\% & 46.70 $\pm$ 0.59 & \cellcolor{green!20}72.88 $\pm$ 0.73 & 86.94 $\pm$ 0.26\% \\
      Idefics2 (one-shot) & 8B & 44.29 $\pm$ 0.62 & 73.84 $\pm$ 0.76 & 90.12 $\pm$ 0.27\% & 39.51 $\pm$ 0.66 & 65.92 $\pm$ 0.80 & 84.03 $\pm$ 0.29\% \\
      InternVL (one-shot) & 8B & \cellcolor{green!20}55.73 $\pm$ 0.50 & \cellcolor{green!20}82.65 $\pm$ 0.66 & \cellcolor{green!20}94.28 $\pm$ 0.22\% & \cellcolor{green!20}49.84 $\pm$ 0.55 & 71.63 $\pm$ 0.70 & \cellcolor{green!20}88.65 $\pm$ 0.24\% \\
      \hline\hline
      \RAE (Ours) (one-shot) & 7B & \cellcolor{green!45}\textbf{62.67 $\pm$ 0.45} & \cellcolor{green!45}\textbf{88.84 $\pm$ 0.53} & \cellcolor{green!45}\textbf{97.67 $\pm$ 0.12}\% & \cellcolor{green!45}\textbf{50.32 $\pm$ 0.49} & \cellcolor{green!45}\textbf{78.10 $\pm$ 0.58} & \cellcolor{green!45}\textbf{89.73 $\pm$ 0.18}\% \\
      \hline
    \end{tabular}
  }
  \caption{One-shot evaluation results of representative LVLMs on our constructed dataset. ``Size'' denotes the number of model parameters in billions. BLEU-4 is reported for Target and Trajectory, and Accuracy for Yes/No. All metrics are in percentages. Our \RAE performs best across tasks.}
  \label{tab:bleu4_acc_table}
\end{table*}

\subsection{Main Comparison}

\textbf{Anomaly Detection Results.}
Table~\ref{tab:val_test_results} summarizes the validation results on UBnormal.
In the one-shot setting, \AHD achieves 94.5\% micro-AUC and 85.2\% macro-AUC, together with 64.3\% RBDC and 74.4\% TBDC, indicating strong localization from a single exemplar.
After fine-tuning, \AHD further improves to 94.8\% micro-AUC and 87.8\% macro-AUC, with 67.8\% RBDC and 76.7\% TBDC, establishing SOTA performance on localization-oriented metrics.
Compared with the strongest baseline (Georgescu~\etal~\cite{georgescu2021background} + UBnormal anomalies), our fine-tuned model gains +26.6 points in micro-AUC (94.8 \vs 68.2), +39.1 in RBDC (67.8 \vs 28.7), and +18.6 in TBDC (76.7 \vs 58.1), while remaining competitive in macro-AUC (87.8 \vs 95.3).
Their macro-AUC advantage relies on a multi-stream architecture, whereas our single-model approach achieves pixel-level anomaly localization while supporting multi-turn dialogue and explainable reasoning.
Similar trends are observed against recent video transformers~\cite{bertasius2021space}, where our method yields substantially higher RBDC/TBDC together with stronger micro-AUC.

\begin{table*}[t!]
  \centering
  \resizebox{\textwidth}{!}{%
    \begin{tabular}{|l|c|cc|cc|c|}
      \hline
      \multirow{2}{*}{\textbf{Method}} &
      \multicolumn{1}{c|}{\textbf{Size $\downarrow$}} &
      \multicolumn{2}{c|}{\textbf{Heatmap (UBnormal) $\uparrow$}} &
      \multicolumn{2}{c|}{\textbf{BLEU-4 (ShanghaiTech) $\uparrow$}} &
      \multicolumn{1}{c|}{\textbf{Acc. (S.T.) $\uparrow$}} \\
      \cline{2-7}
      & \textbf{Parameters} & \textbf{RBDC $\pm$ Std} & \textbf{TBDC $\pm$ Std} & \textbf{Target $\pm$ Std} & \textbf{Trajectory $\pm$ Std} & \textbf{Yes/No $\pm$ Std} \\
      \hline\hline
      \TVAU w/o \AHD & 8299.71M & -- & -- & 61.82 $\pm$ 0.42 & 85.47 $\pm$ 0.51 & 95.38 $\pm$ 0.18\% \\
      \TVAU w/o \RAE & 8317.13M & 67.8 $\pm$ 0.36 & 76.7 $\pm$ 0.41 & -- & -- & -- \\
      \TVAU w/o \AHD \& \RAE & 8274.74M & -- & -- & 61.82 $\pm$ 0.42 & 85.47 $\pm$ 0.51 & 95.38 $\pm$ 0.18\% \\
      \hline\hline
      \TVAU & 8324.67M & 67.8 $\pm$ 0.36 & 76.7 $\pm$ 0.41 & 62.67 $\pm$ 0.45 & 88.84 $\pm$ 0.53 & 97.67 $\pm$ 0.12\% \\
      \hline
    \end{tabular}%
  }
  \caption{Results of different \TVAU variants. ``Parameters/Size'' denotes the number of model parameters in millions. Heatmap metrics include RBDC and TBDC. BLEU-4 is reported for both Target and Trajectory. Accuracy is evaluated on Yes/No classification.}
  \label{tab:tread_ablation}
  \vspace{-4pt}
\end{table*}

\textbf{Multi-turn Dialogue Results.}
To evaluate language understanding and interpretability in multi-turn interactions, we adopt a one-shot setting in which the dialogue follows a hierarchical flow: ``anomaly judgment $\rightarrow$ target appearance attributes $\rightarrow$ motion trajectory/direction $\rightarrow$ temporal anchors and causal cues.'' Compared with representative LVLMs in Table~\ref{tab:bleu4_acc_table}, our \RAE significantly improves textual faithfulness and discriminative power. On ShanghaiTech, BLEU-4 scores for \emph{Target} and \emph{Trajectory} reach \textbf{62.67} and \textbf{88.84}, improving by about 7 and 6 points over the strongest baseline (InternVL2-8B at 55.73/82.65), while Yes/No accuracy increases to \textbf{97.67\%}. On UBnormal, \RAE achieves \textbf{50.32}/\textbf{78.10} BLEU-4 for \emph{Target}/\emph{Trajectory} and \textbf{89.73\%} Yes/No accuracy, again outperforming 7B- and 8B-scale baselines under one-shot prompting.

The performance gain stems not only from the serial coupling of \AHD and \RAE, but also from constructing a dedicated dataset enriched with target-level features, localization, and motion information. This structured dataset supports a curriculum-style SFT process: holistic training on appearance--motion narratives followed by anomaly-focused refinement. By aligning pixel-level heatmaps (\AHD outputs) with structured target timelines and descriptive annotations, \RAE learns to inject region- and time-sensitive prompts into the decoder's semantic space. As a result, the model improves entity disambiguation (``which person/vehicle''), temporal anchoring (``when did the anomaly occur''), and motion semantics (``entering, turning, accelerating''), while reducing hallucinations such as background drift and target switching. Evidence generation, dataset-driven supervision, and prompt injection together establish a closed-loop mechanism for high-fidelity, verifiable multi-turn dialogue.

\subsection{Ablation Studies}

We present a systematic ablation of the two key modules in \TVAU: \AHD and \RAE. These modules are serially coupled: \AHD first generates pixel-level anomaly heatmaps, which are then transformed by \RAE into learnable region-aware prompts injected into the language decoder. Removing both reduces the system to the frozen LVLM backbone, without additional anomaly localization or interpretability.

As shown in Table~\ref{tab:tread_ablation}, the full \TVAU achieves the best performance across all metrics: anomaly localization (RBDC 67.8, TBDC 76.7), text generation quality (BLEU-4 of 62.67 and 88.84), and binary anomaly judgment accuracy (97.67\%), all exceeding the ablated variants.
When \AHD is removed but \RAE is retained, the model loses pixel-level anomaly evidence, making RBDC and TBDC inapplicable. In this case, language outputs rely mainly on the generalization ability of the backbone LVLM, and anomaly descriptions lack spatial interpretability. Conversely, removing \RAE while keeping \AHD still produces high-quality heatmaps (with RBDC/TBDC close to the full model), but without region-aware prompt injection, the language decoder cannot fully exploit the visual evidence, leading to lower BLEU-4 scores and anomaly judgment accuracy.
When both \AHD and \RAE are removed, the system degenerates to the baseline LVLM. Although it can occasionally generate descriptions from large-scale pretraining, overall performance drops sharply: it neither provides reliable anomaly localization nor supports structured anomaly reasoning. This confirms the complementarity and necessity of both modules.

\subsection{Qualitative Analysis}

\begin{figure*}[t!]
  \centering
  \includegraphics[width=1\linewidth]{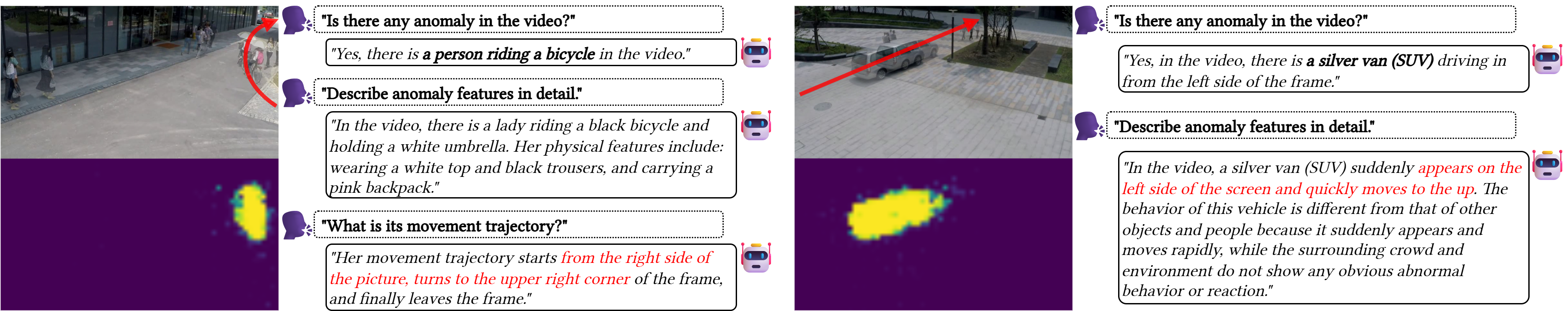}
  \caption{\textbf{Examples of interpretable anomaly detection and multi-turn QA across scenes.}
    Each group shows the raw frame, pixel-level anomaly heatmaps produced by \AHD, and \TVAU's dialogue outputs (anomaly yes/no, appearance/action details, and motion trajectory).
    \textbf{Left:} a cyclist (with umbrella and backpack) is localized as the anomalous target, with the trajectory ``enter from right $\rightarrow$ turn toward the upper-right corner $\rightarrow$ exit.''
    \textbf{Right:} a silver SUV suddenly appears from the left and moves rapidly; \AHD consistently highlights the vehicle, and the QA module explains the abrupt appearance and fast motion.
    \TVAU first detects the anomaly, then describes the appearance (white top, grey shorts, green schoolbag) and the change from walking to running.
  Red arrows indicate the main motion directions, and heatmap intensity reflects anomaly confidence. \RAE encodes the heatmaps into region-aware text prompts that guide the LVLM to produce consistent decisions and descriptions, closing the loop from pixel-level evidence to readable narratives.}
  \label{fig:vis1}
  \label{fig:fig_heatmap}
\end{figure*}

\begin{figure*}[t]
  \centering
  \includegraphics[width=1\linewidth]{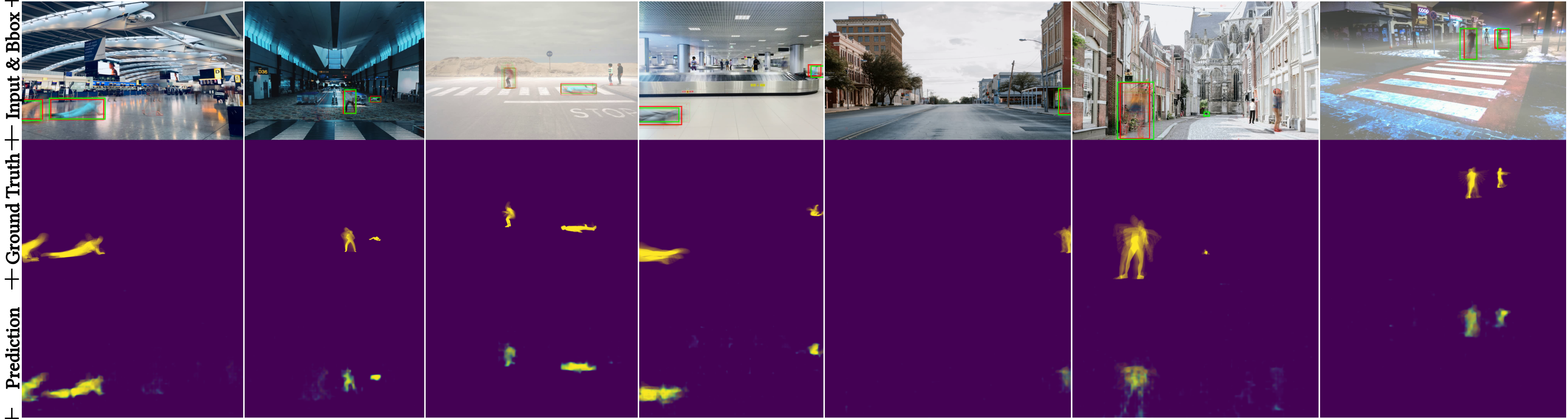}
  \caption{Trajectory visualization by accumulating frame-level outputs. The first row shows multi-frame overlays of the original video with green bounding boxes for GT and red bounding boxes for predictions. The second row overlays GT pixel-level masks to form fine-grained trajectories, while the third row overlays predicted pixel-level masks. Both bounding-box and pixel-level trajectories show strong spatial alignment with GT, indicating that our model accurately captures motion paths over time.}
  \label{fig:vis2}
  \vspace{-4pt}
\end{figure*}

We qualitatively analyze how \TVAU uses heatmaps and region-aware prompts to ensure consistency between pixel-level evidence and language-based reasoning.

Fig.~\ref{fig:fig_heatmap} compares the heatmaps produced by \AHD\ with ground-truth (GT) pixel masks. In pedestrian scenes (\eg, jaywalking and sudden running), \AHD\ concentrates activation on the human silhouette with sharp boundaries while suppressing background structures. In traffic scenes (\eg, abrupt vehicle entry and near-collision), activations adhere to target extents and propagate along the motion direction, producing elongated responses that mirror the trajectory.
Decoder-driven heatmaps (``Video/Image Decoder'' in Fig.~\ref{fig:fig_heatmap}) tend to blur across large regions or drift toward irrelevant textures. In contrast, \AHD\ yields crisper spatial support and tighter overlap with GT masks, especially in crowded scenes.

Fig.~\ref{fig:vis1} illustrates \TVAU in multi-turn interaction. With region-aware prompts from \AHD, the decoder grounds descriptions on the correct spatial regions and time spans. Appearance attributes (\eg, white top, grey shorts, green backpack), motion phrases (\eg, enters from the right, turns upward-right, exits), and state changes (\eg, walking $\rightarrow$ running at frame 10) align with the heatmap. Even with multiple candidates, \RAE\ helps the decoder prioritize the instance supported by the heatmap.
When the anomalous target is partially occluded, \AHD\ responses attenuate but remain anchored and re-amplify upon reappearance. Peak activation follows the target across frames and supports reasoning about entry/exit events, turning points, and acceleration.
Failure cases include micro-actions with minimal displacement, highly nonrigid motion that scatters activation, and scene-dependent appearance shifts (\eg, specularities and fog).
For cross-scene changes and unseen anomaly categories, \AHD\ preserves localization fidelity when target scale is reasonable. For unseen categories sharing motion primitives, \TVAU often highlights the correct target and articulates motion phrases consistent with the evidence.
Heatmaps provide a spatial witness for each claim, enabling justifications such as the silver SUV appearing abruptly from the left and accelerating. Multi-turn follow-ups (which target, ``where'', ``when'') reuse the same evidence chain: \AHD, \RAE, and the decoder. This yields concise, self-consistent narratives and enables error analysis through spatial and temporal evidence.

We further visualize trajectory consistency by overlaying outputs across frames (Fig.~\ref{fig:vis2}). The first row stacks raw frames with bounding boxes (green: GT, red: predictions) for coarse trajectory comparison. The second row accumulates GT pixel-level masks to form fine-grained paths, while the third shows accumulated predicted masks. Both box-level and pixel-level overlays closely match GT, indicating that our method not only localizes anomalies frame by frame but also maintains temporal coherence, producing trajectories consistent with the ground truth over extended periods.

\section{Conclusions}

We present \TVAU, a closed-loop framework that unifies pixel-level anomaly grounding and high-level semantic reasoning by coupling an Anomaly Heatmap Decoder (AHD) with a Region-aware Anomaly Encoder (RAE). By aligning visual features with textual prompts, \TVAU achieves precise, threshold-free spatio-temporal anomaly localization, while its region- and motion-aware prompt design enables LVLMs to perform faithful, structured, and multi-turn anomaly reasoning. This unified formulation goes beyond conventional score-based paradigms, jointly supporting detection, localization, target identification, and explanation within a single framework. Extensive experiments on UBnormal and ShanghaiTech demonstrate consistent improvements over prior methods across localization accuracy, reasoning quality, and dialogue-based evaluation, while ablations confirm the strong complementarity between AHD and RAE.

\section*{Acknowledgments}

We would like to thank Alex Chapiro for insightful discussions and constructive feedback on earlier versions of this manuscript.
The computation was completed on the HPC Platform of the United Arab Emirates University.

\newpage
{
  \small
  \bibliographystyle{ieeenat_fullname}
  \bibliography{main}
}


\end{document}